\documentclass[sigconf]{acmart}

\setcopyright{none}
\renewcommand\footnotetextcopyrightpermission[1]{}

\acmConference[CIKM Applied]{ACM International Conference on Information and Knowledge Management}{Submission Draft}{}

\usepackage{graphicx}
\usepackage{booktabs}
\title{Ishigaki-IDS: An Open-Weight Verifier-Aware Model for Information Delivery Specification Drafting in Building Information Modeling}

\author{Ryo Kanazawa}
\affiliation{%
  \institution{ONESTRUCTION Inc.}
  \city{Tottori}
  \country{Japan}
}
\author{Koyo Hidaka}
\affiliation{%
  \institution{ONESTRUCTION Inc.}
  \city{Tottori}
  \country{Japan}
}
\author{Teppei Miyamoto}
\affiliation{%
  \institution{ONESTRUCTION Inc.}
  \city{Tottori}
  \country{Japan}
}
\author{Takayuki Kato}
\affiliation{%
  \institution{ONESTRUCTION Inc.}
  \city{Tottori}
  \country{Japan}
}
\author{Tomoki Ando}
\affiliation{%
  \institution{ONESTRUCTION Inc.}
  \city{Tottori}
  \country{Japan}
}
\author{Chenguang Wang}
\affiliation{%
  \institution{AWS GenAI Innovation Center}
  \city{Tokyo}
  \country{Japan}
}
\author{Dayuan Jiang}
\affiliation{%
  \institution{AWS GenAI Innovation Center}
  \city{Tokyo}
  \country{Japan}
}
\author{Naofumi Fujita}
\affiliation{%
  \institution{ONESTRUCTION Inc.}
  \city{Tottori}
  \country{Japan}
}
\author{Shuhei Saitoh}
\affiliation{%
  \institution{ONESTRUCTION Inc.}
  \city{Tottori}
  \country{Japan}
}
\author{Atomu Kondo}
\affiliation{%
  \institution{ONESTRUCTION Inc.}
  \city{Tottori}
  \country{Japan}
}
\author{Koki Arakawa}
\affiliation{%
  \institution{ONESTRUCTION Inc.}
  \city{Tottori}
  \country{Japan}
}
\author{Daiho Nishioka}
\affiliation{%
  \institution{ONESTRUCTION Inc.}
  \city{Tottori}
  \country{Japan}
}

\begin{abstract}
Building Information Modeling (BIM) projects require information requirements to be described as machine-checkable Information Delivery Specification (IDS) files in order to verify whether building models contain the required attributes. However, IDS authoring remains a practical bottleneck: practitioners must handle domain vocabulary, strict XML schema constraints, and external validator conformance while also checking whether the requirement itself is correctly expressed. We present Ishigaki-IDS, an open-weight LLM specialized for verifier-aware IDS draft generation. The model combines continued pretraining on BIM/IDS corpora, supervised fine-tuning on information-requirement-to-IDS pairs, and reinforcement learning with verifiable rewards from an external validator. The goal is not to replace expert review, but to move IDS authoring from low-level XML and schema repair toward validator-loadable drafts that practitioners can inspect and correct. On the 166-case expert-created Ishigaki-IDS-Bench, Ishigaki-IDS-8B\footnote{\url{https://huggingface.co/ONESTRUCTION/Ishigaki-IDS-8B}} achieves an IDSAuditPass score of 0.651, a validator-pass metric for generated IDS files, substantially outperforming Claude Opus 4.5, the strongest single-shot LLM baseline we evaluated, at 0.331. It also obtains an Audit-Gated FacetF1 of 0.282, which measures requirement-facet alignment among validator-passing drafts. The same recipe scales: 14B and 32B variants reach IDSAuditPass 0.753 / 0.693 and Audit-Gated FacetF1 0.392 / 0.369. In a workflow check with six BIM practitioners, Ishigaki-assisted authoring reduced aggregate work time by 54.7\% under the same validation and alignment endpoint. These results suggest that verifier-aware IDS generation can reduce the practical burden of converting BIM information requirements into reviewable IDS drafts.
\end{abstract}



\begin{document}
\maketitle

\begin{figure}[!t]
\centering
\includegraphics[width=\linewidth]{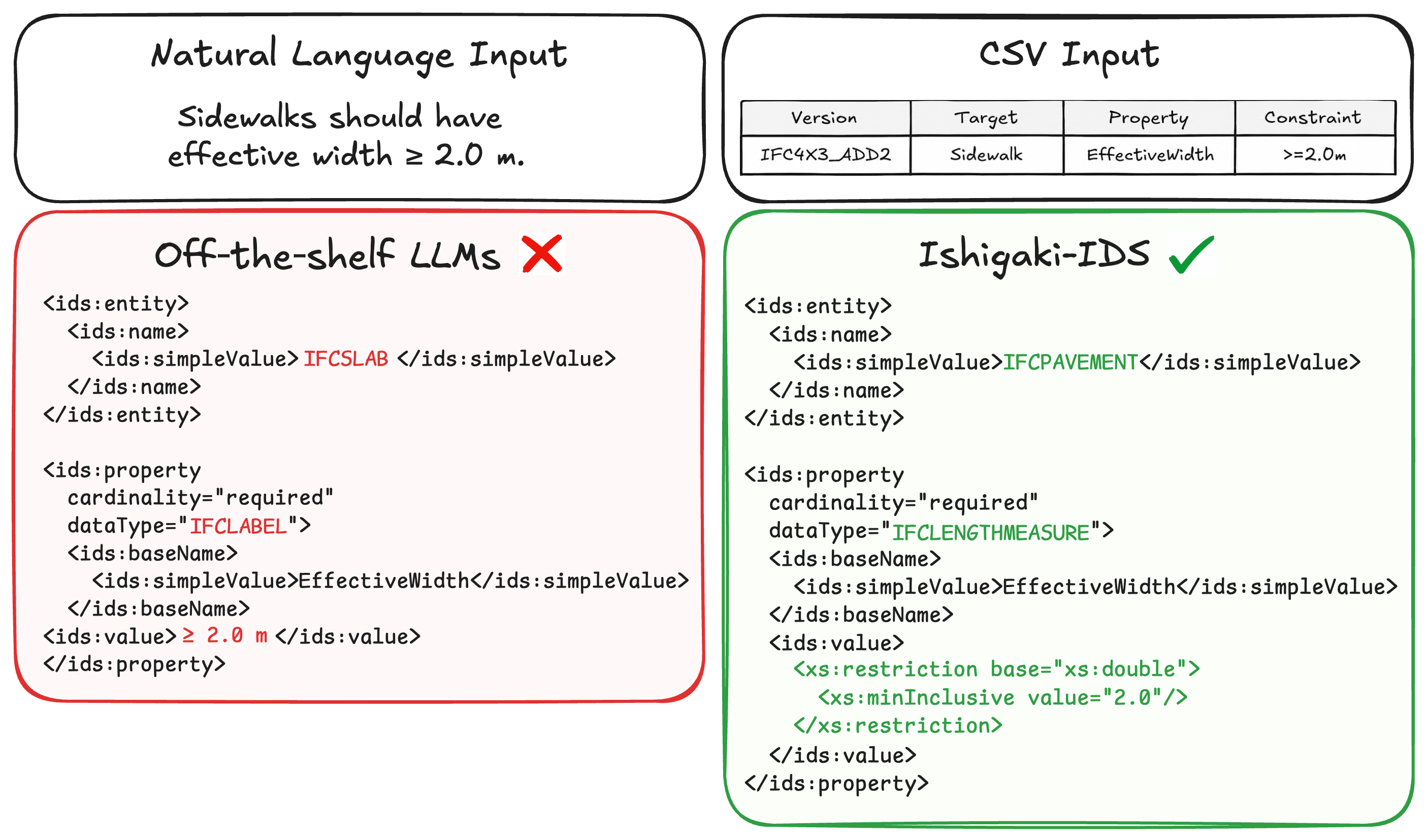}
\caption{Sidewalk IDS generation example, contrasting an incorrect draft with the correct IDS/XML excerpt. The excerpt maps the target to \texttt{IFCPAVEMENT} in IFC4X3\_ADD2 and encodes \texttt{EffectiveWidth} as a required property of datatype \texttt{IFCLENGTHMEASURE} with an \texttt{xs:minInclusive} restriction.}
\Description{Example of converting an effective-width requirement for a sidewalk from natural-language input or CSV input into an IDS/XML fragment. The lower left shows an incorrect IDS draft, and the lower right shows the correct IDS draft using IFCPAVEMENT, IFCLENGTHMEASURE, and xs:minInclusive.}
\label{fig:ids_generation_example}
\end{figure}

\section{Introduction}

Construction projects require information requirements shared across design, construction, and maintenance to be described in a structured form that can be inspected and reused in downstream processes. Such information-requirement authoring can be an application target for structured generation with large language models (LLMs), but in practical domains, producing syntactically valid output alone is insufficient. Generated artifacts must follow domain standards, use standard vocabularies correctly, and be inspectable and usable by external tools used in practice. The Building Information Modeling (BIM) information-requirement authoring problem studied in this work is a representative example of such constrained structured generation. BIM is the concept of a digital model that handles building geometry together with performance and attribute information, and Industry Foundation Classes (IFC) is a data format that represents building elements and attributes handled in BIM with standardized vocabulary and structure\cite{eastman2018bimhandbook,buildingsmart_ifc}. In contrast, Information Delivery Specification (IDS) is not a format for representing a BIM model itself; it is a standard for describing, in an XML schema, which elements in an IFC-represented model should satisfy which property or value conditions\cite{buildingsmart_ids}.

In practice, information requirements are rarely written as IDS from the beginning. They are often provided as specifications, design notes, or tabular checklists. Generating IDS from such inputs requires interpreting the meaning of the requirement sentence, mapping the target building element to the appropriate IFC vocabulary, selecting the required properties and value conditions, and expressing them as an XML structure conforming to the IDS standard. Moreover, unless the generated IDS passes structure and content checks by an external validator such as IDSAuditTool\footnote{\url{https://github.com/buildingSMART/IDS-Audit-tool}}\cite{buildingsmart_idsaudittool}, it cannot be loaded into ordinary validation workflows. Since buildingSMART, an international standards organization, standardized IDS 1.0, the foundation for handling BIM information requirements as machine-readable checking specifications has been taking shape. However, creating an IDS specification requires handling IFC vocabulary, IDS/XML structure, validator interpretation, and project-specific practical requirements at the same time. Therefore, the work of converting specifications or checklists into IDS can become a practical bottleneck in broadly deploying standardized information checking.

This work formulates IDS generation as a structured generation task with verifier constraints and presents Ishigaki-IDS-8B, an open-weight model based on Verifier-Aware Multi-Stage Adaptation. The proposed method combines continued pretraining (CPT) on BIM/IFC/IDS-related corpora, supervised fine-tuning (SFT) on requirement-to-IDS pairs, and reinforcement learning with verifiable rewards (RLVR) using IDSAuditTool. The goal is not to replace expert review, but to generate IDS drafts from specifications or checklists that practitioners can audit and edit, thereby reducing the low-level workload of XML authoring, schema conformance, and mapping to standard vocabulary. On the 166-case expert-created Ishigaki-IDS-Bench, Ishigaki-IDS-8B achieves IDSAuditPass=0.651, substantially outperforming Claude Opus 4.5 at 0.331, the strongest model under the evaluated single-shot prompting setting. We also use Audit-Gated FacetF1 (AG-FacetF1) to jointly evaluate validator passing and alignment with required conditions. Furthermore, in a CSV-to-IDS workflow check with six practitioners, observed work time to reach the same validation and alignment endpoint was reduced by 54.7\%. Ishigaki-IDS-8B has been publicly available on Hugging Face under a CC BY 4.0 license since March 27, 2026.

The contributions of this paper are threefold.
\begin{itemize}
\item We release Ishigaki-IDS-8B~\cite{ishigaki8b_model_card}, an open-weight IDS drafting model that generates validator-checkable drafts from BIM information requirements for practitioner review.
\item We present Verifier-Aware Multi-Stage Adaptation for IDS generation, combining BIM/IFC/IDS continued pretraining, requirement-to-IDS supervised fine-tuning, and RLVR with IDSAuditTool and facet-alignment rewards.
\item We provide applied evidence on Ishigaki-IDS-Bench and a six-practitioner workflow check: Ishigaki-IDS-8B achieves IDSAuditPass 0.651 versus 0.331 for the strongest evaluated single-shot baseline and reduces aggregate time to the fixed endpoint by 54.7\%.
\end{itemize}

\begin{figure*}[!t]
\centering
\includegraphics[width=0.8\textwidth]{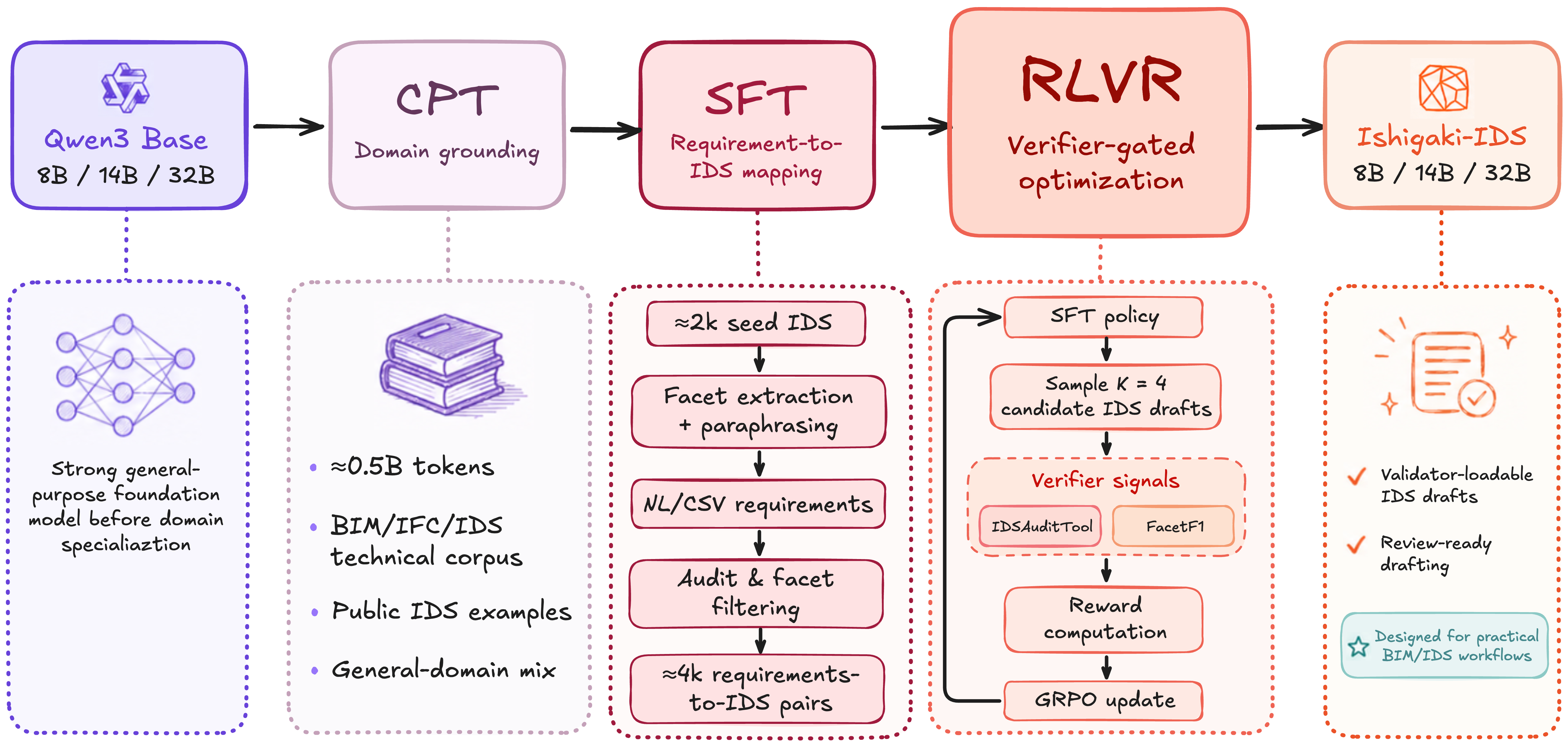}
\caption{
Verifier-Aware Multi-Stage Adaptation of Ishigaki-IDS. CPT adapts the model to BIM/IFC/IDS vocabulary, SFT teaches requirement-to-IDS mapping, and RLVR updates the model with rewards based on IDSAuditTool and FacetF1.
}
\Description{Pipeline diagram in which a Qwen3 base model is adapted into Ishigaki-IDS through CPT, SFT, and RLVR. CPT handles domain knowledge acquisition, SFT handles mapping from requirements to IDS, and RLVR handles verifier-gated optimization.}
\label{fig:pipeline}
\end{figure*}

\section{Related Work}
\label{sec:related_work}

Research on BIM information requirements and IDS has studied how to define information needs as machine-readable checking specifications and connect them to conformance checking over IFC models\cite{eastman2018bimhandbook,buildingsmart_ifc,buildingsmart_ids}. Recent work has investigated support for conversion from tabular requirements to IDS\cite{fischer2025tabularids}, IDS-based BIM model enrichment\cite{demarco2024idsenrichment}, and ontology-based frameworks for IDS evaluation\cite{cerovsek2025ontologyids}, and implementation foundations such as IfcOpenShell and IDSAuditTool have also been developed\cite{ifcopenshell,buildingsmart_idsaudittool}. In the AEC/BIM domain, LLMs have also been applied to code compliance checking, design support, BIM model generation, and information retrieval\cite{madireddy2025llmbimcompliance}; this work focuses on constructing and releasing an open-weight LLM that directly generates IDS drafts capable of passing IDSAuditTool from natural-language or tabular BIM information requirements. IDS generation is also related to semantic parsing\cite{dong2016language,rabinovich2017abstract,yu2018spider}, but it must simultaneously satisfy IFC vocabulary, the IDS standard, external auditing, and facet-level semantic alignment.

Adjacent work on grammar-constrained decoding and structured generation has developed methods that enforce structural validity during generation. Outlines, XGrammar, and vLLM structured outputs provide constrained generation based on regular expressions, JSON schemas, context-free grammars, and related formalisms\cite{willard2023outlines,dong2024xgrammar,vllm_structured_outputs}. These methods can improve XML well-formedness in IDS generation, but they do not solve the problem of combining the building elements, attribute conditions, and value restrictions indicated by the input requirement according to the IDS standard and passing an external validator. Compared with domain-specific LLM studies such as SuperLog\cite{ji2025loganalysis}, Dairy IE\cite{an2025dairyie}, LLP-Bench\cite{brahmbhatt2024llpbench}, and e-commerce catalog studies\cite{luo2025productinfo}, Ishigaki-IDS differs in targeting IDS, a construction standard, and placing validator-loadable artifacts at the center of evaluation.

From the perspective of model adaptation, this work is related to domain adaptation of open-weight LLMs and optimization based on verifiable rewards. LIMA\cite{zhou2023lima}, Tulu 3\cite{lambert2025tulu3}, and Qwen3\cite{yang2025qwen3} provide background for constructing domain-specific models, while PRM800K\cite{lightman2023verify}, DeepSeekMath\cite{shao2024deepseekmath}, and DeepSeek-R1\cite{deepseekai2025r1} show the effectiveness of optimization using verifiable signals. Ishigaki-IDS applies this line of work to IDS authoring, but its reward is composed not of mathematical correctness checking, but of IDSAuditTool structure/content checks and facet alignment against gold IDS.

\section{Task and IDS Representation}
\label{sec:task_system}

Ishigaki-IDS takes as input a BIM information requirement given in natural-language or CSV form, and, when needed, a target IFC version, and outputs a candidate \texttt{.ids} file that practitioners can audit and edit. The task supports the creation of IDS drafts for information requirements applied to IFC models; it is not itself conformance checking of submitted IFC models.

An IDS consists of one or more \texttt{specification} elements. A \texttt{specification} is one checkable rule in an IDS file and has \texttt{applicability}, which defines the target object, and \texttt{requirements}, which defines the conditions that the target must satisfy. In \texttt{applicability}, the target IFC class is typically specified by \texttt{entity}. Here, an IFC class is a category that represents objects in a BIM model, such as walls, floors, pavements, and equipment elements, as standardized vocabulary. In \texttt{requirements}, attributes, properties, datatypes, value restrictions, and other conditions that the target specified in \texttt{applicability} must satisfy are described. For example, a property requirement expresses which information item the target should have, what type its value should have, and what range or enumeration the value should satisfy.

Figure~\ref{fig:ids_generation_example} shows an example of mapping an information requirement given in natural language or CSV to target specifications and condition specifications in IDS/XML. The XML in the figure is not a complete IDS file; it excerpts the core \texttt{ids:entity} specification placed in \texttt{applicability} and the core \texttt{ids:property} specification placed in \texttt{requirements}. As this example shows, IDS generation is not merely XML generation. It is a structured generation task under schema constraints in which the target class, property, datatype, and value restriction corresponding to the input requirement are selected over standard vocabularies.

In this paper, we use \emph{Facet} as a semantically meaningful checking unit in IDS. A Facet is not a unit that mechanically counts single IDS/XML elements one by one, but rather a group of target specifications or condition specifications corresponding to an input requirement. For example, an \texttt{entity} specification in \texttt{applicability} is treated as an entity facet that selects the target IFC class. A property requirement in \texttt{requirements} is treated as a property facet that groups the property name, datatype, required/optional cardinality, and value restriction, including the property-set name when present. The IDS/XML fragment in Figure~\ref{fig:ids_generation_example} shows how an entity facet and a property facet appear in IDS. In evaluation, such Facets are used as comparison units. The detailed canonicalization policy and evaluation metrics are defined in Section~\ref{subsec:metrics}.

\section{Verifier-Aware Multi-Stage Adaptation}
\label{sec:verifier_aware_adaptation}

IDS generation must simultaneously satisfy IDS-standard conformance, IFC vocabulary, and facet-level alignment with the input requirement. We use Qwen3 open-weight LLMs as base models and adapt them to IDS generation through Verifier-Aware Multi-Stage Adaptation, consisting of CPT, SFT, and RLVR. Figure~\ref{fig:pipeline} shows the pipeline applied to 8B, 14B, and 32B models.

\subsection{Adaptation Pipeline}
\label{subsec:adaptation_pipeline}

CPT is the stage that moves the Qwen3-base model closer to BIM/IFC/IDS-domain vocabulary and IDS/XML structure. CPT uses a mixed dataset of about 0.5B tokens. The main sources are a technical corpus containing IFC schemas, IDS specifications, and BIM/IFC/IDS-related documents; public IDS examples collected from public repositories after removing overlap with Ishigaki-IDS-Bench; and a general-domain mix including FineWeb2~\cite{penedo2025fineweb2} to mitigate catastrophic forgetting. The mixing ratio between domain data and general-domain data is approximately 70/30. The objective is to adapt the model distribution to BIM/IFC/IDS terms, IFC entities, IDS rule structure, and XML representation through causal language modeling.

SFT is the stage that learns the mapping from BIM information requirements to IDS drafts. The SFT data is built from about 2k seed IDS files curated from public IFC datasets. Each seed IDS is used after confirming that it passes IDSAuditTool. We then extract a facet set from the seed IDS and paraphrase its meaning into a CSV checklist or natural-language specification while preserving the semantics. For paraphrasing, we use multiple open-weight LLMs whose license conditions were checked, and finally construct about 4k requirement-to-IDS pairs. After synthesis, we filter the data based on facet-set agreement, IDSAuditTool passing of the reconstructed IDS, and XML well-formedness. These paraphrasers are used only as rewriting generators for the SFT data and are not included in the evaluation baselines.

RLVR is the stage that further tunes the post-SFT policy using rewards based on an external verifier and facet alignment. For each input $x$, we sample four candidate IDS drafts and extract IDS fragments from the candidates. We then run IDSAuditTool structure/content checks and compute FacetF1 against the training reference IDS, as defined in Section~\ref{subsec:metrics}. In RLVR, the policy is updated using a verifier-gated reward that combines the IDSAuditTool pass result and this facet-level agreement. For optimization, we use Group Relative Policy Optimization (GRPO) and add a KL constraint with the SFT model as the reference policy. This encourages generation of IDS drafts that pass the validator and preserve the facets corresponding to the input requirement, rather than merely generating IDS-like XML.

\subsection{Training Implementation Details}
\label{subsec:training_details}

CPT, SFT, and RLVR were all run with bf16 mixed precision on a two-node configuration with eight NVIDIA H200 GPUs per node, for a total of 16 GPUs. We used the AdamW optimizer for CPT and SFT optimization. In CPT, the 8B, 14B, and 32B models were continually pretrained for 3 epochs with sequence length 4,096, global batch size 768, and learning rate $1\times10^{-5}$. In SFT, each model was trained for 2 epochs with sequence length 32,768, global batch size 32, and learning rate $5\times10^{-6}$. For CPT and SFT, we used a warmup-stable-decay (WSD) learning-rate scheduler with warmup, stable, and linear-decay ratios of 0.1, 0.8, and 0.1.

In RLVR, each post-SFT model was used as the initial policy and updated with GRPO for 2 epochs. We used the AdamW optimizer for actor updates in GRPO, with training batch size 32, actor learning rate $1\times10^{-6}$, and fixed KL coefficient 0.001. During rollout, for each input, we generated $K=4$ candidate responses with temperature 0.6, top-p 0.95, and top-k 20, using maximum prompt length 3,000 tokens and maximum response length 10,000 tokens. These major settings were shared across the 8B, 14B, and 32B models. The training times from CPT through RLVR for the 8B, 14B, and 32B models were approximately 11.2, 21.1, and 42.7 hours, respectively. CPT and SFT were implemented with NVIDIA NeMo Framework\cite{nemo_framework}, and RLVR was implemented as GRPO on verl\cite{sheng2024hybridflow}. Candidate-response generation during RLVR used vLLM\cite{vllm_structured_outputs}.

\subsection{Verifier-Gated Reward}
\label{subsec:verifier_reward}

Let $z$ be the raw model output and let $y=e(z)$ be the extracted IDS. If IDS extraction fails, $e(z)=\bot$ and the reward is 0. Let $s(y)$ and $c(y)$ denote the structure check and content check of IDSAuditTool, respectively, and let the audit gate $\mathbb{1}_{\mathrm{audit}}(y)=1$ only when both checks pass. With facet-level F1 against gold IDS $y^\ast$ denoted by $F_{\mathrm{facet}}(y,y^\ast)$, the reward is:

\begin{equation}
R(z,y^{\ast}) =
\begin{cases}
0, & e(z)=\bot, \\
\begin{aligned}
&0.1\,s(y)+0.1\,c(y)\\
&\quad+0.8\,\mathbb{1}_{\mathrm{audit}}(y)\,F_{\mathrm{facet}}(y,y^{\ast})
\end{aligned}
& e(z)=y.
\end{cases}
\label{eq:rlvr_reward}
\end{equation}

This design prevents outputs that violate the IDS standard from receiving high reward solely through facet overlap, and places the main signal on facet alignment after validator passing.

\section{Experimental Setup}
\label{sec:experimental_setup}

\subsection{Evaluation Benchmark}
\label{subsec:evaluation_benchmark}

We evaluate IDS generation on Ishigaki-IDS-Bench\cite{kanazawa2026ishigakiidsbench,ishigaki_ids_bench_dataset}\footnote{Code and evaluation materials are available in the Ishigaki-IDS-Bench repository~\cite{ishigaki_ids_bench_software}.}, a curated 166-case evaluation set covering variation in input format, language, target IFC version, and application domain. In this paper, we use Ishigaki-IDS-Bench solely as an evaluation basis; our contribution is Ishigaki-IDS and its verifier-aware adaptation, not benchmark construction. Ishigaki-IDS-Bench is an expert-created benchmark for evaluating IDS generation and contains 108 natural-language inputs and 58 CSV inputs, Japanese and English cases, IFC2X3, IFC4, IFC4X3\_ADD2, and Architectural, Structural, MEP (Mechanical, Electrical, and Plumbing), and General slices. Each sample has an input requirement, target IFC version, and gold IDS for evaluation. Gold IDS files pass the structure/content checks of IDSAuditTool. Ishigaki-IDS-Bench is an evaluation-only dataset and is not used for CPT/SFT/RLVR, prompt tuning, or model selection. We also confirmed that the gold IDS files in Ishigaki-IDS-Bench are not included in the CPT/SFT/RLVR data as exact text matches or file-hash matches.

\subsection{Metrics}
\label{subsec:metrics}

As primary metrics, we measure IDSAuditPass, FacetF1, and AG-FacetF1. IDSAuditPass is the fraction of cases in which the IDS extracted from the generated output passes both the structure check and the content check of IDSAuditTool. IDS extraction failure, XML parse failure, or failure to pass IDSAuditTool is treated as IDSAuditPass=0. All IDSAuditPass results were computed with IDSAuditTool version 1.0.96.

FacetF1 is the F1 between the facet sets extracted from the generated IDS and the gold IDS. Exact string matching would treat XML formatting and semantically unchanged element order as mismatches, so we map facets to canonical tuples containing \texttt{specification}, the \texttt{applicability}/\texttt{requirements} side, facet type, target class, property set, property, datatype, cardinality, value restriction, and related fields before comparison. Differences that can be treated as equivalent under the IDS standard or IFC vocabulary, such as XML formatting, facet order, and value-enumeration order, are absorbed. In contrast, differences that change the meaning or checking result, such as entity, property-set/property, datatype, cardinality, value restriction, letter case in IDS vocabulary, and \texttt{specification} split/merge, are treated as mismatches. IDS extraction failure, XML parse failure, or facet extraction failure gives FacetF1 0. Because target facets and validator results are directly checkable, we do not use LLM-as-judge.

Draft usability is measured by IDSAuditPass, and stricter joint quality is measured by AG-FacetF1. AG-FacetF1 activates FacetF1 only for generated IDS files that pass IDSAuditTool, thereby evaluating generation that satisfies both standard conformance and requirement alignment.
\begin{equation}
\operatorname{AG\mbox{-}FacetF1}(y,y^\ast)
=
\mathbb{1}_{\mathrm{audit}}(y)\cdot
F_{\mathrm{facet}}(y,y^\ast).
\label{eq:audit_gated_facet_f1}
\end{equation}
Here, $y$ is the generated IDS and $y^\ast$ is the gold IDS, and $\mathbb{1}_{\mathrm{audit}}(y)$ is 1 when the generated IDS passes IDSAuditTool and 0 otherwise. FacetF1 alone is reported as a diagnostic measure of how much content corresponding to the input requirement is recovered independently of audit passing.

\subsection{Baselines and Inference}
\label{subsec:baselines_inference}

We evaluate Ishigaki-IDS-8B, Ishigaki-IDS-14B, and Ishigaki-IDS-32B. Ishigaki-IDS-8B is the released open-weight model artifact, and the 14B/32B models are used as scaling variants. External baselines include representative closed proprietary LLMs (GPT-5.5, Claude Opus 4.5, Gemini 3.1 Pro) and recent open-weight LLMs (gpt-oss-120b, Kimi K2.6, DeepSeek V4 Pro, Qwen3.5-397B-A17B). All external baselines were evaluated in May 2026, and the pre-adaptation Qwen3 base models are reported in Table~\ref{tab:ablation}.

All models receive the same zero-shot IDS-generation system prompt. The prompt includes the target IFC version, the input information requirement, and an instruction to output only IDS. If the output contains a fenced code block, the content inside the block is extracted as IDS; otherwise, the whole output is extracted as IDS. IDS extraction failure, XML parse failure, and max-token truncation are treated as invalid outputs, and we do not perform selection from multiple samples, reranking, post-generation repair, regeneration, or manual correction.

Decoding settings are fixed by model family before evaluation and are not tuned on Ishigaki-IDS-Bench. Closed LLMs are evaluated with temperature 0.0, top-p 1.0, and max output tokens 15,000 to the extent accepted by each API. Qwen-family models and Ishigaki variants follow Qwen's public recommended settings: temperature 0.6, top-p 0.95, top-k 20, min-p 0.0, and max output tokens 15,000. Constrained decoding can improve XML well-formedness, but it is not included in the main baselines because this study evaluates single-shot generation ability that simultaneously satisfies IFC vocabulary selection, IDS-standard conformance, validator passing, and facet alignment with the input requirement.

\section{Results and Analysis}
\label{sec:results}

Table~\ref{tab:main_results} reports fixed-run single-shot results on Ishigaki-IDS-Bench. Unless otherwise stated, all values are case averages.

\subsection{Main Results: Validator-Passing IDS Drafts}
\label{subsec:main_results}

\begin{table*}[!t]
\centering
\caption{Main results on Ishigaki-IDS-Bench. IDSAuditPass is the validator-pass rate, FacetF1 is diagnostic requirement alignment without the audit gate, and AG-FacetF1 is the joint quality in Eq.~\eqref{eq:audit_gated_facet_f1}.}
\label{tab:main_results}
\begin{tabular}{llccc}
\toprule
Model group & Model & IDSAuditPass & FacetF1 & AG-FacetF1 \\
\midrule
Closed LLM & GPT-5.5 (xhigh) & 0.277 & \textbf{0.656} & 0.236 \\
Closed LLM & Claude Opus 4.5 & 0.331 & 0.422 & 0.249 \\
Closed LLM & Gemini 3.1 Pro & 0.259 & \underline{0.522} & 0.236 \\
General open-weight LLM & gpt-oss-120b (xhigh) & 0.120 & 0.216 & 0.111 \\
General open-weight LLM & Kimi K2.6 & 0.223 & 0.484 & 0.209 \\
General open-weight LLM & DeepSeek V4 Pro & 0.163 & 0.400 & 0.150 \\
General open-weight LLM & Qwen3.5-397B-A17B & 0.193 & 0.268 & 0.164 \\
Ours (released) & Ishigaki-IDS-8B & 0.651 & 0.359 & 0.282 \\
Ours (scaling) & Ishigaki-IDS-14B & \textbf{0.753} & 0.432 & \textbf{0.392} \\
Ours (scaling) & Ishigaki-IDS-32B & \underline{0.693} & 0.456 & \underline{0.369} \\
\bottomrule
\end{tabular}
\end{table*}

Ishigaki-IDS substantially improves validator-passing IDS draft generation. The released 8B model reaches IDSAuditPass 0.651 (108/166), substantially outperforming Claude Opus 4.5 at 0.331 (55/166), the strongest single-shot proprietary baseline. Its AG-FacetF1 is also 0.282, exceeding Claude Opus 4.5 at 0.249, and it surpasses the best baseline not only in XML/schema conformance but also in joint quality that includes requirement-facet alignment after audit passing.

Table~\ref{tab:main_results} also shows that validator conformance and ungated semantic overlap do not coincide. GPT-5.5 shows FacetF1 0.656, while IDSAuditPass remains 0.277 and AG-FacetF1 is 0.236. Therefore, IDSAuditPass measures the formal endpoint at which a draft can enter a review workflow, and AG-FacetF1 measures the joint quality of standard conformance and requirement alignment.

The 14B scaling variant shows the strongest overall results, reaching IDSAuditPass 0.753 and AG-FacetF1 0.392. The 32B variant raises ungated FacetF1 to 0.456 and obtains AG-FacetF1 0.369. The practical role of Ishigaki-IDS-8B is to convert requirements into validator-loadable drafts often enough for practitioners to review facets inside an IDS workflow, rather than first repairing XML/schema failures. The semantic review burden remaining after validation is analyzed in Section~\ref{subsec:failure_analysis}.

\subsection{Ablation by Adaptation Stage}
\label{subsec:ablation}

\begin{table}[t]
\centering
\setlength{\tabcolsep}{4pt}
\caption{Ablation by adaptation stage on Ishigaki-IDS-Bench. SFT denotes supervised fine-tuning after CPT. Values are based on single fixed-run outputs.}
\label{tab:ablation}
\begin{tabular}{llccc}
\toprule
Size & Stage & IDSAuditPass & FacetF1 & AG-FacetF1 \\
\midrule
8B & Base & 0.157 & 0.202 & 0.133 \\
8B & CPT+SFT & \underline{0.560} & \textbf{0.361} & \underline{0.251} \\
8B & +RLVR & \textbf{0.651} & \underline{0.359} & \textbf{0.282} \\
\midrule
14B & Base & 0.139 & 0.213 & 0.117 \\
14B & CPT+SFT & \underline{0.633} & \underline{0.392} & \underline{0.322} \\
14B & +RLVR & \textbf{0.753} & \textbf{0.432} & \textbf{0.392} \\
\midrule
32B & Base & 0.102 & 0.223 & 0.095 \\
32B & CPT+SFT & \underline{0.627} & \textbf{0.464} & \underline{0.359} \\
32B & +RLVR & \textbf{0.693} & \underline{0.456} & \textbf{0.369} \\
\bottomrule
\end{tabular}
\end{table}

Table~\ref{tab:ablation} reports the adaptation-stage ablation. We compare Base, CPT+SFT, and +RLVR; SFT denotes supervised fine-tuning after CPT. CPT-only checkpoints produced no valid IDS in 0-shot and 1--3-shot trials under the fixed IDS-only protocol, yielding 0 for all three metrics, so we omit them. The results show that the main gain comes from CPT+SFT, and RLVR further improves IDSAuditPass and AG-FacetF1.

\subsection{Practical Scope Across Inputs and Domains}
\label{subsec:slice_analysis}

\begin{table}[t]
\centering
\setlength{\tabcolsep}{3pt}
\caption{Selected slice results for the released Ishigaki-IDS-8B. Slice results are descriptive fixed-run case averages.}
\label{tab:slice_analysis}
\begin{tabular}{llrrr}
\toprule
Axis & Slice & N & IDSAuditPass & AG-FacetF1 \\
\midrule
Input format & CSV & 58 & 0.621 & 0.331 \\
Input format & Natural language & 108 & 0.667 & 0.256 \\
Language & English & 83 & 0.675 & 0.305 \\
Language & Japanese & 83 & 0.627 & 0.259 \\
Domain & Architectural & 48 & 0.688 & 0.223 \\
Domain & Structural & 42 & 0.619 & 0.300 \\
Domain & MEP & 40 & 0.675 & 0.410 \\
Domain & General & 36 & 0.611 & 0.199 \\
\bottomrule
\end{tabular}
\end{table}

Table~\ref{tab:slice_analysis} reports selected slice results for input format, language, and application domain. The audit-pass rate is relatively stable across input formats and languages, while AG-FacetF1 is higher for CSV inputs and MEP and lower for General, where requirement types are more diverse. In other words, Ishigaki-IDS-8B can generate validator-loadable drafts under multiple conditions, but fine-grained alignment with requirement content depends on input format and domain.

\subsection{Failure Analysis and Post-Validation Semantic Residuals}
\label{subsec:failure_analysis}

This subsection analyzes IDS-generation failures from two perspectives. The first is structure-level failure that appears as IDS extraction failure, XML parse failure, or failure to pass the structure check of IDSAuditTool, and the second is facet-level semantic mismatch remaining in outputs that pass IDSAuditTool.

Generic LLM failures are heavily biased toward structure-level failure. The structure-failure rate is 0.592 for closed proprietary LLMs and 0.747 for general open-weight LLMs, compared with 0.124 for Ishigaki-IDS variants. Closed proprietary LLMs and general open-weight LLMs often generate XML fragments or plausible facet candidates, but frequently fail to generate IDS-standard artifacts that pass audit. Ishigaki-IDS reduces this structure-level failure and shifts the main burden of failure toward facet-level semantic mismatches that practitioners should check after validator passing.

\begin{table}[t]
\centering
\caption{Facet agreement among audit-passing outputs from the released 8B model. $N=108$.}
\label{tab:audit_passing_residuals}
\resizebox{0.98\columnwidth}{!}{%
\begin{tabular}{@{}l@{\hspace{0.6em}}r@{\hspace{0.8em}}r@{\hspace{1.0em}}l@{}}
\toprule
FacetF1 bucket & N & \% & Review implication \\
\midrule
1.0 & 14 & 13.0 & Matches the gold facet set under the scorer \\
$[0.75, 1.0)$ & 20 & 18.5 & Minor or local facet checks are needed \\
$[0.5, 0.75)$ & 9 & 8.3 & Several facets require checking \\
$(0, 0.5)$ & 40 & 37.0 & Major facet correction is needed \\
0 & 25 & 23.1 & Passes audit but does not align with gold \\
\bottomrule
\end{tabular}%
}
\end{table}

Table~\ref{tab:audit_passing_residuals} reports results for the 108 outputs of the released 8B model that passed IDSAuditTool. The median FacetF1 among these audit-passing outputs is 0.40, and 65 cases, or 60.2\%, have FacetF1 below 0.5. Therefore, IDSAuditPass should be read not as a metric that guarantees semantic correctness, but as a metric indicating that the output has reached a form that can pass a validator.

Audit-passing examples with FacetF1 0 include a case where the gold IDS requires \texttt{IFCPAVEMENT} for a sidewalk requirement but the model outputs \texttt{IFCSLAB}; a case where room-level conditions are required for dwelling-unit area but the output summarizes them into site-level conditions; and a case where, for floor-slope and slip-resistance requirements, the model generates a non-floor target or an incorrect property. These examples show that Ishigaki-IDS reduces the XML/IDS-standard failures that were common in existing models, but expert checking is still needed to correctly map requirement content to IDS. Thus, the practical value of Ishigaki-IDS lies not in eliminating expert review, but in reducing low-level structural repair and providing drafts that allow practitioners to focus on checking IDS content.

\subsection{Practical Workflow Check with BIM/IDS Practitioners}
\label{subsec:workflow_check}

The previous analysis showed that Ishigaki-IDS makes it easier to generate validator-loadable IDS drafts, while its outputs do not become professionally correct IDS as-is. Therefore, the practical question in this work is not whether expert review can be replaced, but how much the burden of initial IDS construction and XML/schema repair required before review can be reduced. To check this point, we conducted a small exploratory workflow check with six external AEC/BIM practitioners who were neither authors nor internal ONESTRUCTION members. Participants belonged to general construction, design, and MEP-related companies and had roles such as BIM promotion and BIM management, design quality control, technical research, and DX promotion. All six participants had BIM work experience; three had IDS authoring experience, and three did not. Some participants had experience in buildingSMART-related activities, but this exercise was not a buildingSMART validation. No compensation or incentive was provided for participation. Participants consented to the timed exercise and anonymized reporting of aggregate and per-participant timing results; no personally identifiable information is reported.

The task in this experiment was to convert tabular BIM information requirements into IDS. We prepared two corresponding CSV-to-IDS tasks derived from the publicly available Forestry Agency's Wooden BIM Model Parameter Guide\cite{forestry_agency_wooden_bim_parameter_guide}, with comparable input format, target IFC vocabulary, and gold IDS scale. One task handled design and energy-related information requirements, and the other handled envelope area, spatial attributes, dimensions, and identifiers. Each participant completed one task with the OpenAEC\footnote{\url{https://openaec.jp/en}} editor baseline and the other task with the Ishigaki-assisted workflow. Here, OpenAEC was used as a manual IDS editor environment, not as a generative IDS model. In the baseline, participants manually created and corrected IDS in OpenAEC. In the assisted workflow, Ishigaki-IDS-8B first generated an IDS draft from the CSV input, and participants then checked, corrected, validated, and re-corrected the draft in OpenAEC. Regarding OpenAEC, four participants used it for the first time in this exercise, and only two had prior experience using it.

The evaluation endpoint was fixed across the two workflows. The submitted IDS had to pass IDSAuditTool and reach FacetF1=1.0 against a preconstructed gold IDS. During the exercise, participants did not see the gold IDS or facet-level gold annotations. After each submitted candidate, an evaluator ran the fixed completion check and returned only whether the criterion had been met. No gold IDS, missing facet, or target facet information was revealed during the task. The FacetF1 judgment used the same canonicalization policy as Section~\ref{subsec:metrics}, and gold IDS was not shown to participants. Time included the sequence of operations required to satisfy the endpoint.

\begin{table}[t]
\centering
\setlength{\tabcolsep}{4pt}
\caption{Practitioner workflow-check times (s) to the fixed endpoint: IDSAuditPass and FacetF1=1.0; one-sided sign test $p \approx 0.016$.}
\label{tab:practitioner_times}
\begin{tabular}{lrrr}
\toprule
Participant & OpenAEC & Ishigaki-assisted & Reduction \\
\midrule
P1 & 420 & 180 & 57.1\% \\
P2 & 450 & 120 & 73.3\% \\
P3 & 606 & 324 & 46.5\% \\
P4 & 644 & 284 & 55.9\% \\
P5 & 660 & 286 & 56.7\% \\
P6 & 585 & 330 & 43.6\% \\
\midrule
Median & 595.5 & 285.0 & 56.3\% \\
Aggregate & 3365 & 1524 & 54.7\% \\
\bottomrule
\end{tabular}
\end{table}

All six participants satisfied the endpoint faster with Ishigaki assistance (Table~\ref{tab:practitioner_times}). Aggregate time decreased from 3365 seconds for the OpenAEC baseline to 1524 seconds for the Ishigaki-assisted workflow, corresponding to a 54.7\% reduction. Participant-level reductions ranged from 43.6\% to 73.3\%, and median time decreased from 595.5 seconds to 285.0 seconds. However, this result is not evidence of generalized production productivity, and should be positioned as a directional workflow check. In other words, Ishigaki-IDS can reduce the effort required for initial IDS construction and low-level schema repair while retaining expert review.

The released artifact \texttt{ONESTRUCTION/Ishigaki-IDS-8B} has been available on Hugging Face under a CC BY 4.0 license since March 27, 2026\cite{ishigaki8b_model_card}. As of May 20, 2026, it had recorded 98 downloads. In the future, we plan to integrate it into an ONESTRUCTION product and deploy it as a feature that supports IDS draft generation by actual end users.

\section{Limitations and Scope}
\label{sec:limitations}

Ishigaki-IDS-Bench is a benchmark for diagnosing IDS draft-generation ability. It contains curated requirements derived from real-world use cases, but it does not represent all AEC organizations, project types, or IDS authoring practices. In addition, because FacetF1 and AG-FacetF1 depend on the annotation and normalization policies of the benchmark, they do not cover all valid IDS encodings and should be interpreted as policy-based alignment scores.

The main model evaluation uses a zero-shot single-shot generation protocol and, except for the small workflow check with OpenAEC in Section~\ref{subsec:workflow_check}, does not compare against broad end-to-end authoring workflows that combine multiple-sample generation, validator-feedback repair, constrained decoding, retrieval, manual correction, or commercial IDS tools. Also, due to evaluation-budget constraints, we do not report variance across multiple seeds. The practitioner exercise is a controlled check based on six participants and two tasks, so it should be read as directional evidence under the same endpoint, and it does not show that the same effect would hold for other IDS editors, large-scale requirements, workflows spanning multiple organizations, or production use, nor does it demonstrate generalized productivity improvement. Because task assignment and workflow order were not fully counterbalanced, part of the observed work-time reduction may also reflect task familiarity or order effects.

\section{Conclusion}
\label{sec:conclusion}

This work formulated the problem of generating IDS drafts from BIM information requirements as verifier-constrained structured generation that must simultaneously satisfy IDS-standard conformance and facet-level alignment, and constructed and released the open-weight model Ishigaki-IDS through Verifier-Aware Multi-Stage Adaptation. On Ishigaki-IDS-Bench, the released 8B model achieved IDSAuditPass 0.651, exceeding 0.331 for the best non-Ishigaki single-shot baseline, and in a practitioner check with six participants, work time was reduced by 54.7\% under the same endpoint. At the same time, semantic residuals remain after audit passing, so the contribution of this work is not autonomous IDS authoring, but the release of an open-weight IDS draft generator that can hand off drafts to expert review.

\section*{Acknowledgement}

This work was conducted as part of the GENIAC (Generative AI Accelerator Challenge) Project, which aims to strengthen Japan’s capability to develop generative AI and is promoted by the Ministry of Economy, Trade and Industry (METI) and the New Energy and Industrial Technology Development Organization (NEDO).

\section*{GenAI Usage Disclosure}

Generative AI tools were used as part of the research subject and experimental procedures. This includes construction of Ishigaki-IDS variants, synthesis of requirement-to-IDS pairs, and comparative evaluation with baseline models. During manuscript preparation, generative AI was used for language polishing, terminology consistency checking, LaTeX formatting support, and checking Japanese and English expressions. The authors checked the evaluation data, gold IDS, metric computation, and result interpretation, and revised the outputs. The authors are responsible for all claims, experimental design, analysis, conclusions, and the final manuscript.

\bibliographystyle{ACM-Reference-Format}
\bibliography{references}

\end{document}